\documentclass[10pt,twocolumn,letterpaper]{article}

\usepackage[pagenumbers]{iccv} 

\definecolor{iccvblue}{rgb}{0.21,0.49,0.74}
\usepackage[pagebackref,breaklinks,colorlinks,allcolors=iccvblue]{hyperref}
\usepackage{multirow}
\usepackage{makecell}
\usepackage{bm}

\def\paperID{8272} 
\def\confName{ICCV}
\def\confYear{2025}

\title{TRAIL: Transferable Robust Adversarial Images via Latent diffusion}

\author{
Yuhao Xue\textsuperscript{1}\quad\quad
Zhifei Zhang\textsuperscript{1}\quad\quad
Xinyang Jiang\textsuperscript{2}\quad\quad 
Yifei Shen\textsuperscript{2}\quad\quad
Junyao Gao\textsuperscript{1}\\
Wentao Gu\textsuperscript{1}\quad\quad
Jiale Zhao\textsuperscript{1}\quad\quad 
Miaojing Shi\textsuperscript{1}\quad\quad 
Cairong Zhao\textsuperscript{1}\thanks{Corresponding author.}
\\\textsuperscript{1}Tongji University\quad\quad\textsuperscript{2}Microsoft Research Asia\\
\tt\small\{2432200, zhifeizhang, junyaogao, 2331918, 2410917, mshi, zhaocairong\}@tongji.edu.cn,\\ 
\tt\small\{xinyangjiang, yifeishen\}@microsoft.com}

\begin{document}
\maketitle
\begin{abstract}
Adversarial attacks exploiting unrestricted natural perturbations present severe security risks to deep learning systems, yet their transferability across models remains limited due to distribution mismatches between generated adversarial features and real-world data.
While recent works utilize pre-trained diffusion models as adversarial priors, they still encounter challenges due to the distribution shift between the distribution of ideal adversarial samples and the natural image distribution learned by the diffusion model.
To address the challenge, we propose Transferable Robust Adversarial Images via Latent Diffusion (TRAIL), a test-time adaptation framework that enables the model to generate images from a distribution of images with adversarial features and closely resembles the target images. To mitigate the distribution shift, during attacks, TRAIL updates the diffusion U-Net's weights by combining adversarial objectives (to mislead victim models) and perceptual constraints (to preserve image realism). The adapted model then generates adversarial samples through iterative noise injection and denoising guided by these objectives.
Experiments demonstrate that TRAIL significantly outperforms state-of-the-art methods in cross-model attack transferability, validating that distribution-aligned adversarial feature synthesis is critical for practical black-box attacks.
\end{abstract}    
\section{Introduction}
\label{sec:intro}


Adversarial attacks pose a significant threat to modern machine learning models, particularly deep neural networks, by exploiting their vulnerabilities through carefully crafted perturbations. These perturbations, often imperceptible to the human eye, can drastically alter a model’s predictions, leading to severe security risks in critical applications such as autonomous driving, medical imaging, and biometric authentication. 
Unrestricted Adversarial Attack~\cite{hosseini2018semantic, laidlaw2019functionaladversarialattacks, bhattad2019unrestricted, yuan2022natural, chen2024content, chen2024diffusion} go beyond conventional adversarial attack paradigms by replacing strict constraints on perturbation magnitude with unrestricted
 but natural changes on images. These methods modify the image's content~\cite{chen2024content}, color~\cite{hosseini2018semantic, zhao2020adversarial}, object shape~\cite{xiao2018spatially}, texture~\cite{bhattad2019unrestricted}, etc., altering the image while preserving as much semantic coherence with the target image as possible, making detection and defense significantly more challenging. 

While unrestricted attacks is able to add more complex adversarial features to the target image, they still face significant challenges in transferability—meaning the ability of an adversarial example generated for one model to remain effective against other models with different architectures, training data, or decision boundaries. 
Certain types of perturbations, such as style or texture modifications, may not consistently alter decision boundaries across different classifiers, limiting their effectiveness in a transfer-based attack scenario. This raises a critical question of what makes an adversarial attack transferable. 

A previous theoretical study \cite{allen2022feature} suggests that an adversarial perturbation is more likely to be transferable if it consists of a dense mixture of robust image features. However, most conventional optimization-based adversarial attacks lack mechanisms to enforce this constraint. As a result, their transferability depends on whether the attack features learned from the source model are also present in the target model, limiting their effectiveness across different architectures. 
Recently, large text-to-image generative models trained on extensive datasets have demonstrated an impressive ability to generate realistic images that closely match real-world data distributions \cite{rombach2022high, radford2021learning, peebles2023scalable, nichol2021glide, saharia2022photorealistic}, demonstrating a remarkable ability to generate realistic images that closely align with real-world data distributions. This advancement creates new opportunities for improving attack transferability. By leveraging generative models as a prior distribution for adversarial features, attack methods are more likely to produce perturbations that incorporate real-world image features, increasing their effectiveness across different models. 

Some pioneering works \cite{chen2024content, chen2024diffusion} have proposed a straightforward approach that incorporates pre-trained diffusion models into adversarial attacks as a prior constraint by utilizing them to generate adversarial samples. 
Instead of directly perturbing the original image, this method embeds adversarial information into the target image’s latent representation after the diffusion forward process, then reconstructs it into image space using a pre-trained generative model. 
However, this approach still faces challenges. First, there is a distribution mismatch between the generated adversarial images, which follow the distribution learned by the pre-trained diffusion model, and the ideal adversarial samples, which should retain the natural image distribution while incorporating a dense mixture of robust adversarial features. This distribution shift may cause the obtained adversarial sample lack the necessary perturbations to effectively fool the target model. Second, diffusion model-based attacks may achieve high attack performance at the cost of  introducing more notable changes to the target image than conventional methods, due to the inherent randomness in the noise sampling process and the adversarial information added to the latent embeddings. 

In this paper, we propose to address these challenges with a novel test-time adaptation (TTA)-based approach that dynamically updates the pre-trained diffusion model using the target images during the attack, called (T)ransferable (R)obust (A)dversarial (I)mages via (L)atent diffusion ({\bf TRAIL}). 
This adaptation enables the model to generate images from a distribution of real-world images with adversarial features and closely resembles the target image. 
Specifically, given a target image, TRAIL first updates the weights of the U-Net in the pre-trained diffusion model using an objective loss that encourages the model to generate adversarial samples capable of altering the target model’s predictions while maintaining high similarity to the original image. 
After TTA, TRAIL generates adversarial samples by first adding noise to the input image and then progressively denoising it through the adapted diffusion model, guided by adversarial gradients. 
We conducted extensive experiments, demonstrating that TRAIL achieves significant improvement on attack performance and transferability. 

In summary, we highlight our contributions as follows:

\begin{itemize}
\item We propose a novel attack framework, TRAIL, the first test-time adaptation framework for adversarial attacks that enables the pre-trained diffusion model to generate adversarial images from real-world distribution, which have better attack transferability.

\item We investigate the characteristics of adversarial attacks that improve transferability across different model architectures and analyze how the effectiveness of various attack methods is affected when applied to models with different architectures.

\item We perform extensive experiments and demonstrate that TRAIL significantly outperforms existing methods in terms of black-box attack success rates. Additionally, to the best of our knowledge, we are the first to explore attack transferability on vision-language models trained with large-scale data.

\end{itemize}

\section{Related Works}
\label{sec:2}


\subsection{Adversarial Attacks}
Adversarial attacks against deep neural networks (DNNs) are broadly categorized into restricted and unrestricted attacks based on perturbation constraints. Restricted attacks generate adversarial examples via imperceptible, norm-bounded perturbations. Szegedy et al. \cite{szegedy2013intriguing} first demonstrated that small perturbations could induce misclassification and transfer across models. Goodfellow et al. \cite{goodfellow2014explaining} linked this vulnerability to DNN linearity, proposing the Fast Gradient Sign Method (FGSM) for efficient attacks. Madry et al. \cite{madry2017towards} formalized robustness via robust optimization, defining security against first-order adversaries. AdvGAN \cite{xiao2018generating} further accelerated attack generation using GANs. 
Unrestricted attacks \cite{qiu2020semanticadv, jia2022adv, bhattad2019unrestricted} manipulate semantic features to create realistic adversarial examples. Semantic attacks \cite{hosseini2018semantic} alter hue/saturation in HSV space, preserving human-recognizable semantics while reducing model accuracy. Functional attacks (e.g., ReColorAdv \cite{laidlaw2019functionaladversarialattacks}) apply global color/texture shifts, combining with restricted perturbations to bypass defenses. Bhattad et al. These attacks prioritize semantic coherence over perturbation size, challenging defenses reliant on noise detection. However, these methods are highly effective in white-box attacks but perform poorly in black-box scenarios.

\subsection{Transferable Attacks}
Transferable adversarial attacks \cite{xiong2022stochastic, li2024transferable, yang2024quantization} aim to craft adversarial examples that remain effective across unknown models, a critical capability for real-world scenarios where attackers lack access to target model architectures or defense mechanisms.
In restricted attacks, Momentum-based methods \cite{dong2018boosting} stabilize gradient updates to enhance transferability. Admix \cite{wang2021admix} augments inputs by mixing them with diverse samples, broadening adversarial generalization. Variance tuning \cite{wang2021enhancing} optimizes gradients using historical variance, while frequency-domain augmentation \cite{long2022frequency} diversifies surrogate models via spectral transformations. Flat local maxima optimization \cite{ge2023boosting} prioritizes flat loss regions to improve cross-model robustness. In unrestricted Attacks, Content-based methods \cite{chen2024content} leverage generative models (e.g., Stable Diffusion~\cite{rombach2022high}) to craft adversarial examples aligned with natural image manifolds. DiffAttack \cite{chen2024diffusion} employs diffusion models to generate imperceptible perturbations in latent spaces, targeting both model predictions and human perception.
These works highlight gradient stabilization for restricted attacks and generative models for unrestricted attacks, challenging defenses through enhanced cross-model transferability. However, despite focusing on transferability, these methods produce perturbations that remain highly dependent on the surrogate model, restricting their effectiveness across different target models.
\section{Methodology}

In this section, we elaborate the detail of TRAIL from four aspects. 
In Section \ref{subsec:motivations}, we discuss the adversarial attack transferability motivated from a theoretical perspective. 
In section \ref{subsec:diff-tta}, we introduce the test-time adaptation of pre-trained diffusion model for adversarial attack. 
Section \ref{subsec:diff-generation} elaborates on the adversarial gradient guided sampling process that generates adversarial images. 
In Section \ref{subsec:diffusion_fine_tuning}, we elaborate the one-step gradient propagation method used in the adaptation process.

\subsection{What Makes an Adversarial Attack Transferable?}
\label{subsec:motivations}
We explore the characteristics of adversarial attacks that enhance transferability across different model architectures, drawing insights from a theoretical perspective. A rigorous study (Section 6 of \cite{allen2022feature}) identifies the dense mixture of robust features as a key factor in generating transferable adversarial perturbations. Specifically, a small perturbation aligned with the average direction of these robust features tends to transfer effectively between models. Building on this foundation, we analyze the extent to which different adversarial attack methods maintain their effectiveness when applied to models with varying architectures.

\paragraph{Optimization-based attack: Why is it transferable?} 
We begin by analyzing classic optimization-based adversarial attacks. Let \( n \) different classifiers, after applying the softmax function, be represented as \( p_{\phi_1}(y|\bm{x}), \dots, p_{\phi_n}(y|\bm{x}) \), where \( p_{\phi_i}(y|\bm{x}) \) denotes the predicted probability of class \( y \) given input \( \bm{x} \) for the \( i \)th classifier.  
Suppose we craft an adversarial perturbation using the first classifier, aiming to misclassify the input \( \bm{x} \) as a target class \( y_1 \). The objective of the optimization-based attack is formulated as: 
\begin{equation}
\begin{aligned}
    \min_{\bm{\delta}}  -\log p_{\phi_1}(y_1 | \bm{x} + \bm{\delta}) \quad \text{subject to} \quad \|\bm{\delta}\| \leq \epsilon,
\end{aligned}
\label{eq:opt}
\end{equation}
where \( \bm{\delta} \) is the adversarial perturbation, constrained by a norm bound \( \epsilon \) to ensure the perturbation remains small. This formulation encourages the model to assign a higher probability to the incorrect class \( y_1 \), thereby maximizing the likelihood of mis-classification.

Using a Lagrange multiplier \( \lambda \), the constrained optimization problem can be reformulated as:  
\begin{equation}
\begin{aligned}
    \min_{\bm{\delta}} -\log p_{\phi_1}(y_1|\bm{x}+\bm{\delta}) + \lambda \|\bm{\delta}\|.
\end{aligned}  
\label{eq:reformulated}
\end{equation}
This can be rewritten as:  
\begin{equation}
\begin{aligned}\label{eq:rewritten}
\min_{\bm{\delta}}
    -\log p_{\phi_1}(y_1|\bm{x}+\bm{\delta}) - \log \underbrace{\exp(-\lambda \|\bm{\delta}\|)}_{p(\bm{x} + \bm{\delta})}.
\end{aligned}  
\end{equation}
This formulation implies that the perturbation \( \bm{\delta} \) follows a prior distribution proportional to \( \exp(-\lambda \|\bm{\delta}\|) \), which is independent of the image features. Consequently, the optimization-based attack primarily exploits the adversarial features present in the classifier itself to generate perturbations.  
As a result, the attack is transferable to another classifier \( p_2 \) only if \( p_1 \) and \( p_2 \) share the same adversarial features for the given input \( \bm{x} \) and target class \( y_1 \).

\paragraph{Transferability of generation-based attack. } 
The analysis highlights a key limitation of conventional optimization-based adversarial attacks: their reliance on the adversarial features learned by the source classifier. This limitation suggests the need for a more generalizable approach to adversarial attacks—one that does not depend solely on the source model’s feature space. A promising solution is to introduce a generative prior as $p(x + \bm{\delta})$, leveraging a pre-trained generative model to guide the adversarial attack. By incorporating such a prior, the generated adversarial perturbations are more likely to align with robust and universal image features. 

\paragraph{Tackling distribution shift and constraining the distance.} 
A straightforward approach utilizes diffusion models to project the target image into the latent space, where a perturbation $\bm{\delta}$ is applied to the latent representation $ \bm{x} $ to improve attack transferability. This ensures that the adversarial image, when reconstructed by the diffusion model, follows the natural image distribution learned by the pre-trained generative model, making the attack more effective. 
However, this method faces several challenges.  
The first challenge is the \emph{distribution shift} between the natural image \( \bm{x} \) and the adversarial image \( \bm{x} + \bm{\delta} \). While diffusion models are designed to generate images that strictly follow the natural image distribution (i.e. $p(x)$), adversarial examples should instead follow a distribution that includes a dense mixture of robust features in addition to natural image characteristics (i.e. $p(x+\delta)$ where $\delta$ contains robust image features). To address this issue, we adopt test-time adaptation (TTA) \cite{liang2025comprehensive}, which dynamically updates the weights of the diffusion model to better align with the adversarial objective. 
The second challenge is the large transformation distance between the generated adversarial image and the original input. Due to the iterative nature of diffusion models, the adversarial image may deviate significantly from \( \bm{x} \), potentially making the attack perceptible. To mitigate this, we introduce a constraint in TTA that explicitly limits the distance between the generated adversarial image and the original image.  

\paragraph{Why choose diffusion models?} 
As discussed above, a transferable attack relies on a dense mixture of robust features. Therefore, it is crucial to choose a generation model that inherently captures robust representations. Diffusion models have been shown to exhibit certified robustness, whereas other generation models like GANs lack such guarantees \cite{chen2025diffusion}. Consequently, diffusion models are the preferred choice for generating transferable adversarial examples.



\begin{figure*}[t]
  \centering
  \includegraphics[width=1\linewidth]{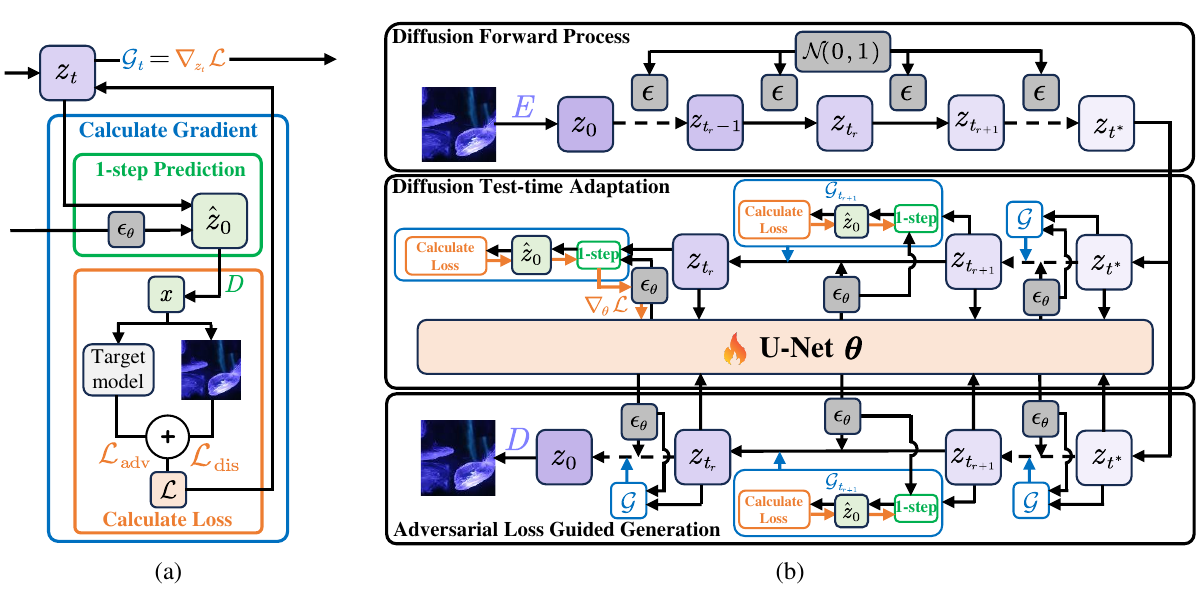} 
   \caption{Overview of our approach. (a) We adapt the diffusion model with the optimization goal of generating more effective adversarial images while minimizing modifications to the original image. We use this objective to guide the adaptation and generation processes, enhancing the attack effectiveness of the generated images. (b) We present the complete process of generating an adversarial image. We employ test-time adaptation (TTA) by updating the diffusion model’s weights to better align with our adversarial objective. We introduce guidance in denoising process to enhance the attack effective of our generated image (Section \ref{subsec:diff-generation}). During TTA, we perform backpropagation for adaptation in a single prediction step. (Section \ref{subsec:diffusion_fine_tuning}). }
   \label{method_figure}
\end{figure*}

\subsection{Diffusion Model Test-time Adaptation}
\label{subsec:diff-tta}
We first formulate the problem of Adversarial Attacks.
Given an image-label pair $(x, y)$, where $x\in\mathcal{X}$ is an original image from the training set used to train the target classifier and $y\in\mathcal{Y}$ is the ground truth label of $x$, a target model $f:\mathbb{R}^n\mapsto [K]$ where $K$ is the number of classes. 
An adversarial image $\hat{x}$ crafted by $x$ should satisfy:
\begin{equation}
{\underset{i\in[K]}{\operatorname{arg max}}} f(\hat{x})\neq y,
\label{eq:problem_definition}
\end{equation}

To generate such image, we adopt test-time adaptation (TTA) to the diffusion model (DM). In each iteration, DM takes $x$ as input, generates the adversarial image $\text{DM}(x)$, and updates the model’s weights by optimizing two loss functions:
\textbf{Adversarial loss}, we choose negative cross-entropy loss as the adversarial loss $\mathcal{L}_\text{adv}$ which mainly guides $\text{DM}(x)$ toward misclassification: 
\begin{equation}
\mathcal{L}_{adv}(\text{DM}(x),y)=-\sum_{i=1}^K\mathbb{I}(i=y)\log p_{\phi}(i|\text{DM}(x)),
\label{eq:ladv_fomula}
\end{equation}
where $p_{\phi}$ is the probability for the ith-class predicted by the classifier $\phi$, and \textbf{Distance loss}: we choose mean squared error loss as the distance loss $\mathcal{L}_\text{dis}$ which mainly guides $\text{DM}(x)$ to be as close as possible to $x$:
\begin{equation}
\mathcal{L}_{dis}(\text{DM}(x),x)=\frac{1}{N}\|x-\text{DM}(x)\|_F,
\label{eq:ldis_fomula}
\end{equation}
where $N$ is the total number of pixels in $x$, $\|\cdot\|_F$ represents the Frobenius norm.
In general, the objective function is as follows, where $\alpha,\beta$ represent the weight factors of each loss:
\begin{equation}
\min_\theta \mathcal{L}(x,\text{DM}(x),y)=\alpha\mathcal{L}_\text{adv}(\text{DM}(x),y)+\beta\mathcal{L}_\text{dis}(\text{DM}(x),x),
\label{eq:objective_function}
\end{equation}

\subsection{Adversarial Loss guided Diffusion Generation}
\label{subsec:diff-generation}

In this section, given a target image, we describe how to use a diffusion model to generate a corresponding adversarial sample using aderverial loss as guidance. During the forward process of diffusion, Gaussian noise is progressively added to the latent image $z_0$. \cite{li2023adaptive} have demonstrated that the information of the original target image is gradually reduced as well. 
As a result, when the image is reconstructed using the diffusion denoiser, its features are altered. 
By fine-tuning the diffusion model with Eq.\ref{eq:objective_function} introduced in the previous section, we ensure that these alterations embed adversarial information. 
Specifically,  at each denoising step, the U-Net denoiser is able to inject adversarial perturbations into $z_t$, effectively guiding the generation of adversarial samples.
Thus, the adversarial sample can be generated by first adding gaussian noise to the target image for $t^*$ steps and then reconstructing the noisy image with the test-time adapted diffusion model.  

Specifically, given a target image $x$, we first use VAE encoder to map it to latent image $z_0$, then gradually add Gaussian noise and obtain the final noise image $z_{t^*}$:
\begin{equation}
q(z_t|z_{t-1})\sim\mathcal{N}(\sqrt{1-\beta_t}z_{t-1},\beta_t\mathbf{I}),
\label{eq:forward_process}
\end{equation}
where pre-defined $\beta_t\in (0,1)$ is a sequence of positive noise scales. 
For each image generation, we use the diffusion model to denoise $z_{t^*}$ over $t^*$ steps, ultimately obtaining the noise-free latent variable $z_0$:
\begin{equation}
p_\theta(z_{t-1}|z_t)\sim\mathcal{N}(\mu_\theta(z_t,t),\Sigma_\theta(z_t,t)),
\label{eq:denoising_process}
\end{equation}
where $\mu_\theta(x_t,t)$ is predicted by a pre-trained network $\epsilon_\theta(x_t,t)$:
\begin{equation}
\mu_\theta(z_t,t)=\frac{1}{\alpha_t}(x_t-\frac{\beta_t}{\sqrt{1-\bar{\alpha}_t}}\epsilon_\theta(x_t,t)),
\label{eq:mu_theta}
\end{equation}
$\alpha_t=1-\beta_t, \bar{\alpha}_t=\prod_{i=1}^t \alpha_i$ and $\Sigma_\theta(x_t,t)$ is a constant depending on $\beta_t$. Finally, we apply the VAE decoder to reconstruct the final image $\text{DM}(x)$.
However, there is a trade-off between attack performance and stealthiness depending on the choice of $t^*$. A larger $t^*$ introduces stronger adversarial perturbations, enhancing attack effectiveness, while a smaller $t^*$ preserves closer distance between the generated and original images.

\textbf{Proposition}. \textit{Let $z_0(t^*)$ be the denoised latent variable generated from the noisy latent $z_{t^*}$ via the reverse SDE process. Assume the normalized noise prediction error satisfies $||-\frac{1}{\sqrt{1-\bar{\alpha}_t}}\epsilon_\theta(z_t,t)||_2^2\leq C$ for all $z_t$, $t\in[0,1]$, and let $\delta\in(0,1)$. Then, with probability at least $(1-\delta)$, }
\begin{equation}
\begin{aligned}
&\|z-z_0(t^*)\|_2^2 \\ 
\leq \ & \sigma^2(t^*)(C\sigma^2(t^*)+d_z+2\sqrt{-d_z\log\delta}-2\log\delta),
\end{aligned}
\label{eq:proposition}
\end{equation}
\textit{where $z=E(x)$ is the VAE-encoded latent, $z_0(t^*)$ is the denoising result from $z_{t^*}$ and $d_z$ is the latent dimension of $z$.}

We provide the proof in Appendix. To generate more effective adversarial examples after fine-tuning, $t^*$ needs to be larger. However, according to the proposition above, as $t^*$ increases, the distance between the obtained $z_0$ and the original $z$ also become larger, causing the generated $\text{DM}(x)$ to deviate significantly from the original image $x$.

To generate adversarial images through the diffusion process, we incorporate gradient guidance into the denoising steps, leveraging classifier signals to directly influence the model’s denoising decisions toward adversarial outputs.
At denoising step $t$, we use the following equation to predict the denoised $z_0$ from the current $z_t$ using $\epsilon_\theta$:
\begin{equation}
p_\theta(\hat{z}_0|z_t)  \sim \ \mathcal{N}(\frac{1}{\sqrt{\bar{\alpha}_t}}z_t-\sqrt{\frac{1-\bar{\alpha}_t}{\bar{\alpha}_t}}\epsilon_\theta(z_t,t),\Sigma_\theta(z_t,t)).
\label{eq:one_step_prediction}
\end{equation}
Mapping $\hat{z}_0$ back to the image space, we obtain the predicted adversarial image $\hat{x}_0$. We then compute the gradient w.r.t. $z_t$ using the objective function in Eq.\ref{eq:objective_function} to derive the guidance $\mathcal{G}$.
We use $\mathcal{G}_t$ to guide diffusion model during sampling in each steps:
\begin{equation}
p_\theta(z_{t-1}|z_t)\sim\mathcal{N}(\mu_\theta(z_t,t)+\Sigma_\theta(z_t,t)\mathcal{G}_t,\Sigma_\theta(z_t,t)),
\label{eq:denoising_with_guidance}
\end{equation}
The guidance is introduced in both the test-time adaptation and final generation denoising processes. We use this guidance to direct the diffusion model's generation, resulting in adversarial images with enhanced attack effectiveness.

\subsection{One-step Back Propagation for Diffusion TTA}
\label{subsec:diffusion_fine_tuning}
During the test-time adaptation process, the gradient originates from the objective function $\mathcal{L}(x, \text{DM}(x), y)$ and flows towards the diffusion model at the generated image $\text{DM}(x)$. It then passes through every simpling steps in the denoising process to reach $\theta$. 
However, we observe that directly propagating the gradient in this way remains challenging: 1) The generation process involves inherent randomness, and as the number of samples increases, the gradients become increasingly unstable. 2) Introducing guidance into the generation process increases the complexity of the computation graph, significantly raising memory requirements and demanding more hardware resources for gradient backpropagation.

To address these challenges, inspired by \cite{xu2023imagereward}, we minimize the sampling process along the gradient backpropagation path as much as possible. Specifically, we randomly select an intermediate timestep $t_r$. During adaptation, in each iteration we perform the denoising process up to timestep $t_r$ without gradients and obtain $z_{t_r}$. Then, we directly predict $z_0$ from $z_{t_r}$ with $\epsilon_\theta$ by Eq.\ref{eq:one_step_prediction}. Similar to the process in calculating the guidance, we decode $\hat{z}_0$ into $\hat{x}_0$ to compute $\mathcal{L}$, and then perform backpropagation to obtain $\partial \mathcal{L} / \partial \theta$. In this way, we update $\theta$ with a small computational cost. 

\section{Experiments}
\label{sec:4}



\begin{table*}[t]
\centering
\scalebox{0.8}{
\begin{tabular}{ccccccccccc}
\toprule
\multirow{2}{*}{\begin{tabular}[c]{@{}c@{}}Surrogate\\ Model\end{tabular}} & \multirow{2}{*}{Method} & \multicolumn{8}{c}{Attacked Models}                                                                                               & \multirow{2}{*}{\begin{tabular}[c]{@{}c@{}}Transfer\\ Avg \%\end{tabular}} \\ \cline{3-10}
                                                                           &                         & Mn-v2          & Inc-v3        & RN-50          & Den-161       & RN-152        & EF-b7         & Vit-b-32       & Swin-B         &                                                                            \\ \midrule
-                                                                          & original                & 12.1           & 19.9          & 7.0            & 5.7           & 5.7           & 14.3          & 11.1           & 3.8            &                                                                            \\ \midrule
\multirow{8}{*}{MobileNet-v2}                                              & SAE                     & 92.2*          & 32.2          & 35.8           & 29.1          & 27.8          & 30.1          & 31.3           & 15.9           & 28.6                                                                       \\
                                                                           & ReColorAdv              & 97.8*          & 33.4          & 32.4           & 23.2          & 23.7          & 25.7          & 24.4           & 14.7           & 25.4                                                                       \\
                                                                           & cAdv                    & 95.9*          & 38.2          & 33.4           & 27.3          & 26.0          & 35.6          & 33.2           & 19.0           & 30.4                                                                       \\
                                                                           & ColorFool               & 65.0*          & 28.7          & 25.0           & 16.8          & 17.1          & 21.9          & 21.6           & 6.1            & 19.6                                                                       \\
                                                                           & NCF                     & 93.3*          & 48.5          & 66.3           & 42.3          & 55.6          & 41.9          & 49.4           & 24.1           & 46.9                                                                       \\
                                                                           & ACA                     & 93.9*          & 66.5          & 61.1           & 56.6          & 58.1          & 58.9          & 56.8           & 48.9           & 58.1                                                                       \\
                                                                           & DiffAttack              & 98.1*          & 69.1          & 75.8           & 66.2          & 66.4          & 61.4          & 44.2           & 53.1           & 62.3                                                                       \\
                                                                           & ours                    & \textbf{99.5*} & \textbf{81.3} & \textbf{90.5}  & \textbf{80.4} & \textbf{81.7} & \textbf{75.3} & \textbf{52.2}  & \textbf{65.7}  & \textbf{75.3}                                                              \\ \midrule
\multirow{8}{*}{ResNet-50}                                                 & SAE                     & 48.7           & 30.9          & 87.0*          & 30.5          & 28.2          & 30.1          & 34.0           & 17.3           & 31.4                                                                       \\
                                                                           & ReColorAdv              & 40.3           & 31.4          & 96.2*          & 28.7          & 33.0          & 25.9          & 22.7           & 14.6           & 28.1                                                                       \\
                                                                           & cAdv                    & 41.7           & 36.4          & 97.6*          & 33.0          & 31.9          & 35.7          & 35.1           & 20.1           & 33.4                                                                       \\
                                                                           & ColorFool               & 41.6           & 30.3          & 67.3*          & 19.6          & 21.6          & 23.9          & 24.2           & 7.1            & 24.0                                                                       \\
                                                                           & NCF                     & 70.4           & 48.7          & 90.3*          & 48.2          & 59.7          & 39.9          & 50.3           & 25.0           & 48.9                                                                       \\
                                                                           & ACA                     & 71.3           & 68.9          & 90.0*          & 62.0          & 65.3          & 62.3          & 59.3           & 53.8           & 63.3                                                                       \\
                                                                           & DiffAttack              & 77.0           & 70.6          & 96.6*          & 78.9          & 80.7          & 63.4          & 44.9           & 55.6           & 67.3                                                                       \\
                                                                           & ours                    & \textbf{87.3}  & \textbf{78.0} & \textbf{99.0*} & \textbf{85.7} & \textbf{86.7} & \textbf{72.4} & \textbf{48.4}  & \textbf{67.1}  & \textbf{75.1}                                                              \\ \midrule
\multirow{8}{*}{Vit-b-32}                                                  & SAE                     & 42.3           & 31.3          & 33.8           & 27.2          & 27.2          & 30.7          & 83.7*          & 16.9           & 29.9                                                                       \\
                                                                           & ReColorAdv              & 35.2           & 31.7          & 25.2           & 19.4          & 21.2          & 27.6          & 94.5*          & 15.3           & 25.1                                                                       \\
                                                                           & cAdv                    & 37.3           & 38.0          & 29.2           & 25.7          & 24.7          & 35.9          & 95.9*          & 22.5           & 30.5                                                                       \\
                                                                           & ColorFool               & 38.9           & 34.0          & 32.9           & 21.3          & 24.0          & 26.2          & 70.8*          & 9.5            & 26.7                                                                       \\
                                                                           & NCF                     & 60.6           & 45.6          & 56.0           & 38.4          & 49.1          & 40.2          & 85.5*          & 27.7           & 45.4                                                                       \\
                                                                           & ACA                     & 65.0           & 69.0          & 59.1           & 58.4          & 59.5          & 62.7          & 90.8*          & 53.9           & 61.1                                                                       \\
                                                                           & DiffAttack              & 61.7           & 67.0          & 61.2           & 58.7          & 59.9          & 64.4          & 99.5*          & \textbf{58.8}  & 61.7                                                                       \\
                                                                           & ours                    & \textbf{73.4}  & \textbf{72.8} & \textbf{68.8}  & \textbf{61.1} & \textbf{63.3} & \textbf{68.5} & \textbf{99.9*} & 52.8           & \textbf{65.8}                                                              \\ \bottomrule
\end{tabular}
}
\caption{Performance comparison of adversarial transferability on normally trained CNNs and ViTs. The adversarial examples are generated by MobileNet-v2, ResNet-50, Vit-b-32, and SwinB, respectively. "*" indicates white-box attack settings.}
\label{table_unrestricted}
\end{table*}

\begin{table*}[h]
\centering
\scalebox{0.8}{
\begin{tabular}{ccccccccccc}
\toprule
           & $\text{Inc-v3}_\text{ens3}$ & $\text{Inc-v3}_\text{ens4}$ & $\text{IncRes-v2}_\text{ens}$ & R\&P          & JPEG          & Hu            & NRP           & NRP-resG      & SID           & DiffPure      \\ \midrule
SAE        & 11.5          & 12.0          & 10.9            & 21.3          & 14.0          & 16.6          & 14.8          & 15.0          & 26.5          & 55.3          \\
ReColorAdv & 13.4          & 14.6          & 12.9            & 11.5          & 17.3          & 17.0          & 16.1          & 17.7          & 25.8          & 58.9          \\
cAdv       & 13.0          & 17.1          & 10.4            & 24.3          & 11.1          & 26.4          & 21.9          & 20.6          & 34.0          & 65.0          \\
ColorFool  & 20.6          & 21.5          & 19.3            & 30.4          & 19.6          & 31.5          & 26.0          & 29.1          & 29.2          & 66.8          \\
NCF        & 31.3          & 33.0          & 26.4            & 40.3          & 28.7          & 43.9          & 37.3          & 39.5          & 40.9          & 68.7          \\
ACA        & 58.6          & 61.9          & 56.0            & 54.6          & 42.0          & 56.1          & 49.4          & 51.5          & 61.5          & 70.4          \\
DiffAttack & 60.1          & 63.2          & 57.1            & 56.0          & 47.5          & 60.2          & 48.5          & 52.0          & 61.0          & 72.1          \\
ours       & \textbf{64.5} & \textbf{66.0} & \textbf{62.3}   & \textbf{61.2} & \textbf{56.8} & \textbf{66.9} & \textbf{53.7} & \textbf{57.1} & \textbf{69.8} & \textbf{77.4} \\ \bottomrule
\end{tabular}
}
\caption{The ASR (\%) against several defenses. The adversarial examples are crafted by attacking MobileNet-v2.}
\label{table_defense}
\end{table*}

\subsection{Experimental Setup}
\label{subsec:4-1}

\paragraph{Dataset.} Our experiments are conducted on the ImageNet compatible dataset, including 1,000 images of resolution $299\times 299\times 3$ from the ImageNet validation set. This dataset has been widely used in previous works~\cite{dong2018boosting, gao2020patch}.

\paragraph{Models.} To evaluate the white-box attack capabilities of our method and its transferability in black-box settings, we selected two types of network architectures for our experiments: Convolutional Neural Networks (CNNs) and Vision Transformers (ViTs). We assessed the performance of our method on models with varying architectures. For CNNs, we chose MobileNet-V2~\cite{sandler2018mobilenetv2}, Inception-v3~\cite{szegedy2016rethinking}, DenseNet-161~\cite{huang2017densely}, ResNet-50 and ResNet-152~\cite{he2016deep}, and EfficientNet-B7~\cite{tan2019efficientnet} as the target models. For ViTs, we selected ViT-b-32~\cite{dosovitskiy2020image} as the target models. We also selcet Swin-B~\cite{liu2021swin} which is shown in the appendix.

\paragraph{Implementation Details.} Our experiments are run on NVIDIA GeForce RTX 3090 with Pytorch. The following parameters are used in experiments: U-Net fine-tune number of iterations $N=100$, Diffusion steps $T=80$,  $t^*=8$, and a learning rate of $10^{-5}$. We use Stable Diffusion v2.0~\cite{rombach2022high} as DM.

\begin{figure*}[t]
  \centering
  \scalebox{0.9}{
    \includegraphics[width=1\linewidth]{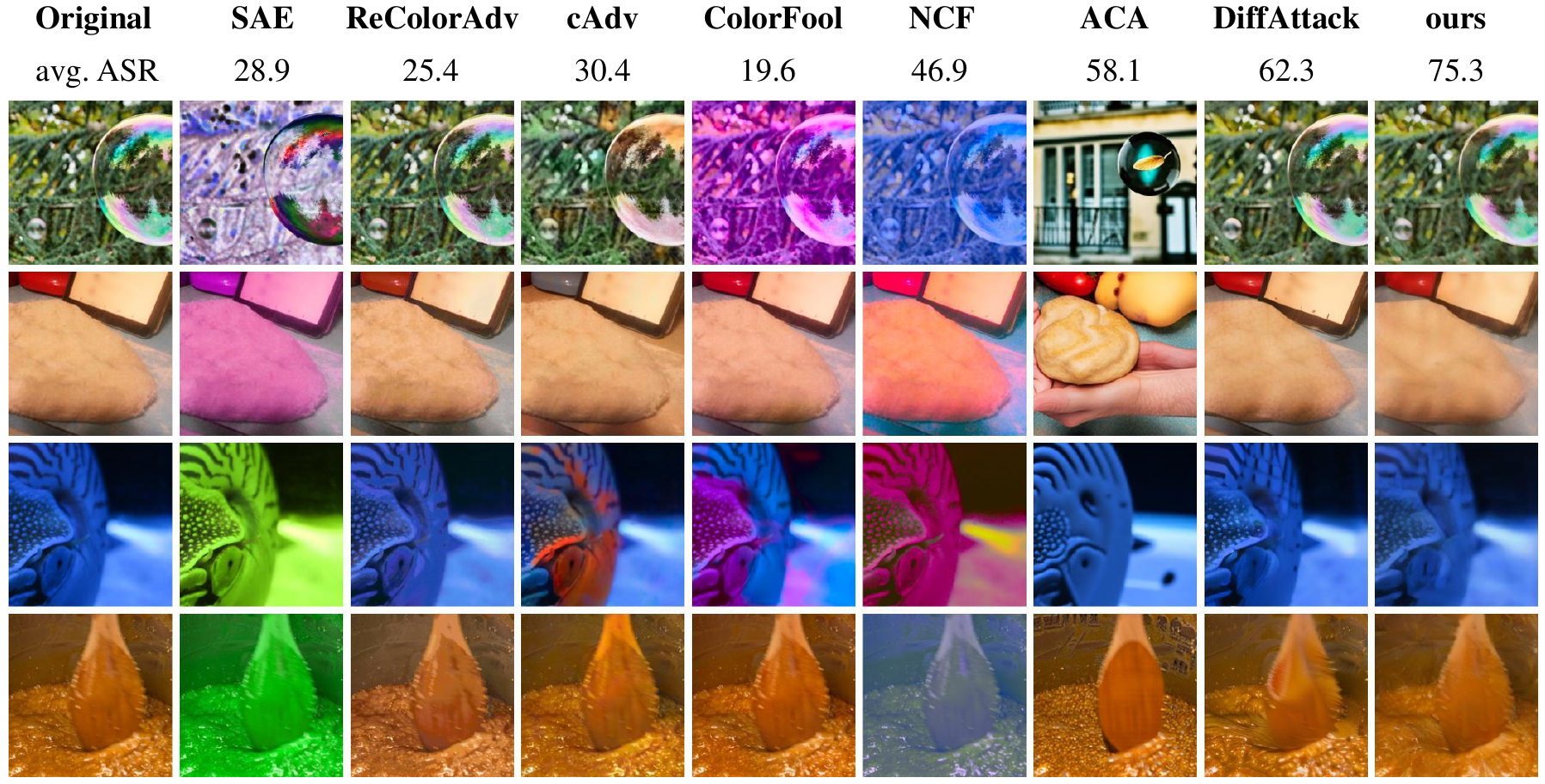} 
  }
   \caption{Compared to other methods, our approach produces more effective adversarial images with minimal alterations to the original images. The avg. ASR (\%) represents the average attack success rate in a black-box setting. A higher value indicates stronger attack transferability.
    }
   \label{example_show_figure}
\end{figure*}

\subsection{Unrestrict Adversarial Attack Comparisons}
\label{subsec:4-2}
In this section, we evaluate the transferability of unrestricted attacks on normally trained convolutional neural networks (CNNs) and vision transformers(ViTs).
We selected SAE~\cite{hosseini2018semantic}, ReColorAdv~\cite{laidlaw2019functionaladversarialattacks}, cAdv~\cite{bhattad2019unrestricted}, ColorFool~\cite{shamsabadi2020colorfool}, NCF~\cite{yuan2022natural}, ACA~\cite{chen2024content}, DiffAttack~\cite{chen2024diffusion} for comparison with our method. 

Table \ref{table_unrestricted} presents a comparison of the attack performance of unrestricted adversarial attack methods under both white-box and black-box settings. The "Transfer Avg\%" in the last column indicates the average attack success rate across the seven black-box attack settings, excluding the white-box attack setting. Compared to other state-of-the-art unrestricted adversarial attack methods, our method achieves a significant lead in the attack transferability performance of the generated adversarial examples. Our method performs exceptionally well in attacking both CNN and ViT models. Notably, ACA and DiffAttack are also diffusion-based attack methods, and our method achieves superior performance without requiring more time than these diffusion-based methods. Compared to the current state-of-the-art unrestricted attack, DiffAttack, our method achieves at least $7.8\%$ and $4.1\%$ higher performance when using CNNs and ViTs as surrogate models, respectively.

Figure \ref{example_show_figure} compares adversarial images generated by our method and other prior works. 
Our method achieves a significant improvement in attack transferability while making minimal modifications to the image, aligning with the goals of unrestricted attacks.

\subsection{Attacks on Defense Methods}
\label{subsec:4-4}
While many attack methods can easily mislead classification models, strong defense methods now exist that process the image to nullify the attack, enabling the model to correctly classify it again. Many defense techniques are now effective at detecting and removing $l_p$-norm perturbations, achieving good defense results. To further demonstrate the advantages of TRAIL, we selected R\&P~\cite{xie2017mitigating}, JPEG~\cite{guo2017countering}, Hu~\cite{hu2019new}, NRP~\cite{naseer2020self}, NRP-resG~\cite{naseer2020self}, SID~\cite{tian2021detecting}, and DiffPure~\cite{nie2022diffusion} as defense methods to defend against adversarial examples generated by various attack methods, including TRAIL, chose MobileNet-v2 as the target model, and compared the defense robustness of these attack methods, as shown in Table \ref{table_defense}. This edge in performance is present when using both CNN and ViT as surrogate models. It can be observed that our method effectively bypasses current defense methods and demonstrates superior defense robustness compared to all existing restricted adversarial attack methods by a large margin on average. Especially under the defense of the current most powerful purification method, DiffPure, we still maintain a $77.4\%$ attack success rate. This underscores that current adversarial attack methods remain destructive against well-designed defense techniques, highlighting an urgent need in the security field to enhance the robustness of adversarial defenses.

\begin{figure}[t]
\scalebox{0.9}{
  \includegraphics[width=1\linewidth]{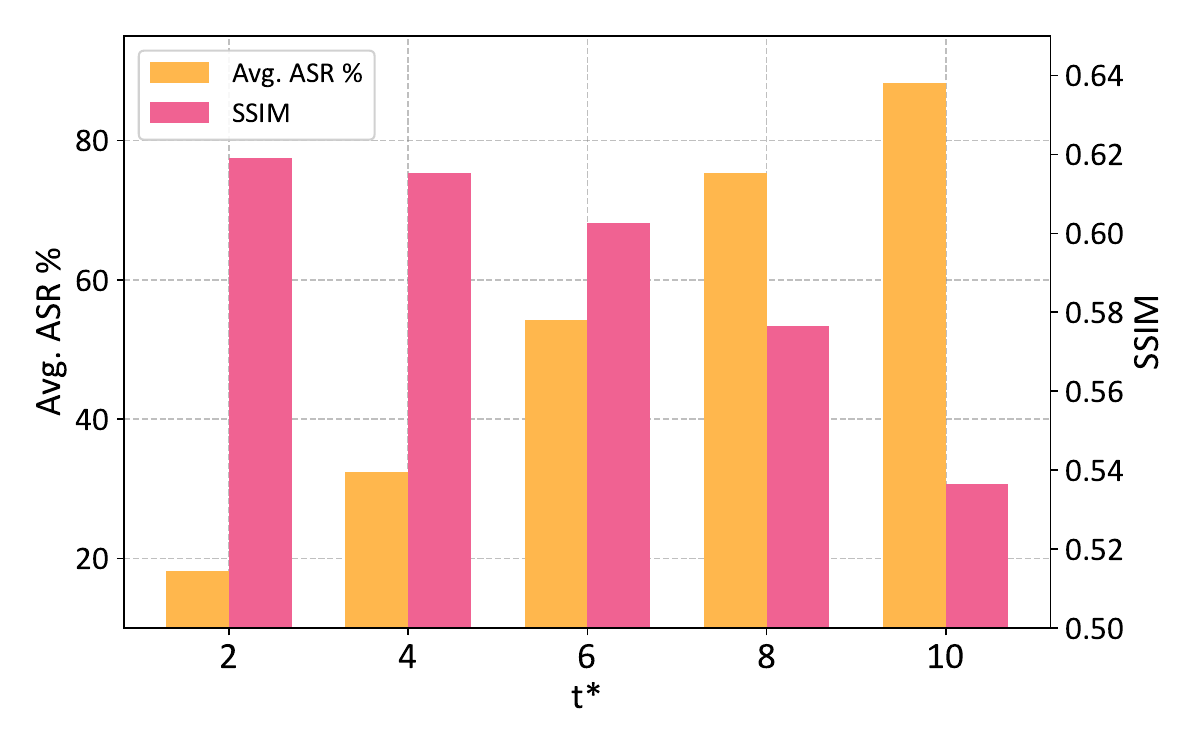} 
  }
   \caption{Trade-off between attack performance and stealthiness depending on the choice of $t^*$.}
   \label{fig:trade_off_figure}
\end{figure}

\subsection{Trade-off on $t^*$}
\label{subsec:exp_trade_off}
As mentioned in Section \ref{subsec:diff-generation}, there is a trade-off between attack performance and stealthiness depending on the choice of $t^*$. We use MobileNet-v2 as the surrogate model and conduct attacks at different $t^*$ (we select $t^*$ to be 2,4,6,8,10) while keeping all other variables unchanged. We evaluated the effect of different $t^*$ values on the attack success rate (ASR) against MobileNet-v2 and the structural similarity index (SSIM)~\cite{wang2004image} between the generated adversarial examples and the original images. SSIM measures the similarity between two images, with higher values indicating greater resemblance. As shown in Figure \ref{fig:trade_off_figure}, as $t^*$ increases, our method introduces more adversarial information into the latent image by the adaptated diffusion model during generation, resulting in a higher attack success rate. Conversely, a higher $t^*$ increases the number of sampling steps during generation, amplifying the difference between the adversarial and original images and reducing the SSIM value.

\subsection{Black-box Attack on Vision Language Model}
\label{subsec:exp_clip_and_llava}
Recent advancements in visual foundation models (e.g., CLIP~\cite{radford2021learning}, DINOv2~\cite{oquab2023dinov2}) and multimodal large language models (e.g., GPT-4V, LLaVA~\cite{liu2023llava}) have significantly enhanced AI's ability to perform generalized and robust vision understanding. By training on large-scale vision datasets, these models enable more effective cross-modal reasoning, zero-shot learning, and adaptable visual perception across diverse tasks.
However, despite their robustness, we demonstrate that TRAIL poses a significant threat to these advanced AI models, even in a black-box setting, using a small CNN as a surrogate model, revealing vulnerabilities even in extensively pre-trained models. 

\begin{figure}[t]
\centering
\scalebox{0.8}{
  \includegraphics[width=1\linewidth]{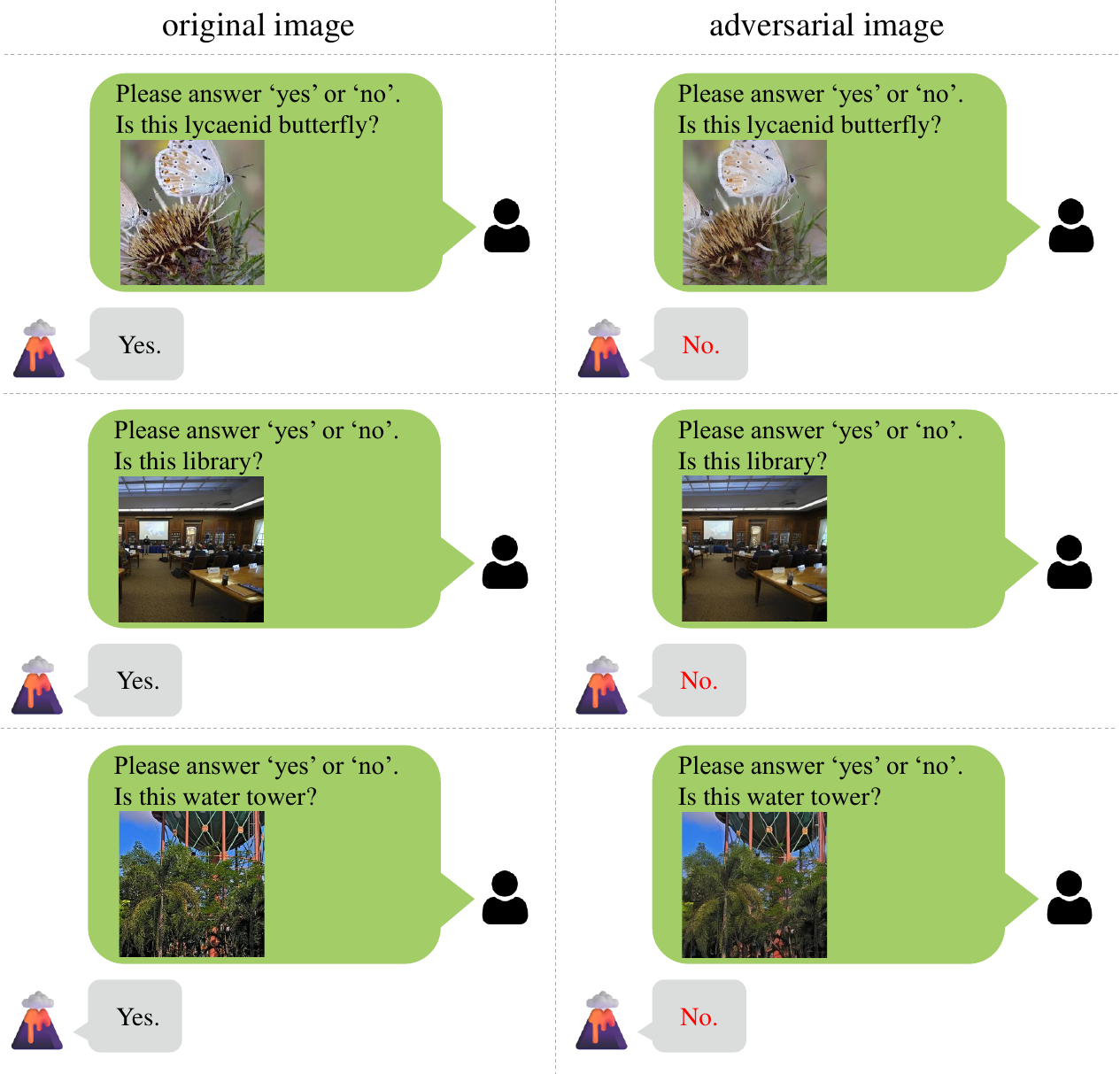} 
  }
   \caption{The evaluation process on LLaVA. In the left column, LLaVA evaluates original images and correctly associates them with their labels. In the right column, LLaVA analyzes adversarial images generated by TRAIL on the surrogate model and is misled into rejecting the correct label.}
   \label{fig:llava_evaluating_figure}
\end{figure}

We select clip-vit-large-patch14 and LLaVA-1.5-7B as target models, select MobileNet-v2 and ResNet-50 as surrogate model. For CLIP, we provide 1,000 text labels, and it calculates similarity scores between the test image and each label. The label with the highest score is selected as the final prediction. For LLaVA, we use a question-answering approach, evaluating each image as an image-prompt to the model and processing the responses, as shown in Figure \ref{fig:llava_evaluating_figure}. Due to the token limit, we cannot include all category names in the prompt for LLaVA to select the most similar one. Instead, we verify whether the image belongs to its correct category. Table \ref{table:clip_llava_table} presents a comparison of the attack performance of unrestricted adversarial attack methods under black-box settings to CLIP and LLaVA, and TRAIL still outperforms other methods. However, all methods exhibit limited attack effectiveness on LLaVA, highlighting the need for further research.

\begin{table}[]
\centering
\scalebox{0.7}{
\begin{tabular}{ccccc}
\toprule
Surrogate Model & \multicolumn{2}{c}{MobileNet-v2} & \multicolumn{2}{c}{ResNet-50} \\
Attack Method   & CLIP14          & LLaVA          & CLIP14        & LLaVA         \\ \midrule
Original        & 30.6            & 6.9            & 30.6          & 6.9           \\ \midrule
SAE             & 39.7            & 8.3            & 37.0          & 7.7           \\
ReColorAdv      & 34.8            & 7.1            & 34.6          & 7.3           \\
cAdv            & 39.2            & 8.3            & 39.6          & 8.6           \\
ColorFool       & 34.2            & 6.5            & 33.9          & 6.4           \\
NCF             & 38.2            & 7.0            & 38.1          & 7.6           \\
ACA             & 55.0            & 20.9           & 56.3          & 22.0          \\
DiffAttack      & 54.1            & 12.9           & 55.9          & 15.6          \\
ours            & \textbf{69.0}   & \textbf{24.0}  & \textbf{70.4} & \textbf{25.1} \\ \bottomrule
\end{tabular}
}
\caption{The ASR (\%) on clip-vit-large-patch14 and LLaVA-1.5-7B.}
\label{table:clip_llava_table}
\end{table}

\section{Conclusion}

In this paper, we introduce TRAIL, the first test-time adaptation framework for adversarial attacks. TRAIL enables a pre-trained diffusion model to generate adversarial images that closely follow real-world distributions while achieving superior attack transferability.
We explore the key factors that enhance adversarial attack transferability across different model architectures and analyze how various attack methods perform when applied to models with different structures.
Extensive experiments demonstrate that TRAIL significantly outperforms existing methods in black-box attack success rates.
{
    \small
    \bibliographystyle{ieeenat_fullname}
    \bibliography{main}

\begin{thebibliography}{54}
\providecommand{\natexlab}[1]{#1}
\providecommand{\url}[1]{\texttt{#1}}
\expandafter\ifx\csname urlstyle\endcsname\relax
  \providecommand{\doi}[1]{doi: #1}\else
  \providecommand{\doi}{doi: \begingroup \urlstyle{rm}\Url}\fi

\bibitem[Allen-Zhu and Li(2022)]{allen2022feature}
Zeyuan Allen-Zhu and Yuanzhi Li.
\newblock Feature purification: How adversarial training performs robust deep learning.
\newblock In \emph{2021 IEEE 62nd Annual Symposium on Foundations of Computer Science (FOCS)}, pages 977--988. IEEE, 2022.

\bibitem[Bhattad et~al.(2019)Bhattad, Chong, Liang, Li, and Forsyth]{bhattad2019unrestricted}
Anand Bhattad, Min~Jin Chong, Kaizhao Liang, Bo Li, and David~A Forsyth.
\newblock Unrestricted adversarial examples via semantic manipulation.
\newblock \emph{arXiv preprint arXiv:1904.06347}, 2019.

\bibitem[Chen et~al.(2025)Chen, Dong, Shao, Zhongkai, Yang, Su, and Zhu]{chen2025diffusion}
Huanran Chen, Yinpeng Dong, Shitong Shao, Hao Zhongkai, Xiao Yang, Hang Su, and Jun Zhu.
\newblock Diffusion models are certifiably robust classifiers.
\newblock \emph{Advances in Neural Information Processing Systems}, 37:\penalty0 50062--50097, 2025.

\bibitem[Chen et~al.(2024{\natexlab{a}})Chen, Chen, Chen, Zhang, Zou, and Shi]{chen2024diffusion}
Jianqi Chen, Hao Chen, Keyan Chen, Yilan Zhang, Zhengxia Zou, and Zhenwei Shi.
\newblock Diffusion models for imperceptible and transferable adversarial attack.
\newblock \emph{IEEE Transactions on Pattern Analysis and Machine Intelligence}, 2024{\natexlab{a}}.

\bibitem[Chen et~al.(2024{\natexlab{b}})Chen, Li, Wu, Jiang, Ding, and Zhang]{chen2024content}
Zhaoyu Chen, Bo Li, Shuang Wu, Kaixun Jiang, Shouhong Ding, and Wenqiang Zhang.
\newblock Content-based unrestricted adversarial attack.
\newblock \emph{Advances in Neural Information Processing Systems}, 36, 2024{\natexlab{b}}.

\bibitem[Dong et~al.(2018)Dong, Liao, Pang, Su, Zhu, Hu, and Li]{dong2018boosting}
Yinpeng Dong, Fangzhou Liao, Tianyu Pang, Hang Su, Jun Zhu, Xiaolin Hu, and Jianguo Li.
\newblock Boosting adversarial attacks with momentum.
\newblock In \emph{Proceedings of the IEEE conference on computer vision and pattern recognition}, pages 9185--9193, 2018.

\bibitem[Dosovitskiy(2020)]{dosovitskiy2020image}
Alexey Dosovitskiy.
\newblock An image is worth 16x16 words: Transformers for image recognition at scale.
\newblock \emph{arXiv preprint arXiv:2010.11929}, 2020.

\bibitem[Gao et~al.(2020)Gao, Zhang, Song, Liu, and Shen]{gao2020patch}
Lianli Gao, Qilong Zhang, Jingkuan Song, Xianglong Liu, and Heng~Tao Shen.
\newblock Patch-wise attack for fooling deep neural network.
\newblock In \emph{Computer Vision--ECCV 2020: 16th European Conference, Glasgow, UK, August 23--28, 2020, Proceedings, Part XXVIII 16}, pages 307--322. Springer, 2020.

\bibitem[Ge et~al.(2023)Ge, Liu, Xiaosen, Shang, and Liu]{ge2023boosting}
Zhijin Ge, Hongying Liu, Wang Xiaosen, Fanhua Shang, and Yuanyuan Liu.
\newblock Boosting adversarial transferability by achieving flat local maxima.
\newblock \emph{Advances in Neural Information Processing Systems}, 36:\penalty0 70141--70161, 2023.

\bibitem[Goodfellow et~al.(2014)Goodfellow, Shlens, and Szegedy]{goodfellow2014explaining}
Ian~J Goodfellow, Jonathon Shlens, and Christian Szegedy.
\newblock Explaining and harnessing adversarial examples.
\newblock \emph{arXiv preprint arXiv:1412.6572}, 2014.

\bibitem[Guo et~al.(2017)Guo, Rana, Cisse, and Van Der~Maaten]{guo2017countering}
Chuan Guo, Mayank Rana, Moustapha Cisse, and Laurens Van Der~Maaten.
\newblock Countering adversarial images using input transformations.
\newblock \emph{arXiv preprint arXiv:1711.00117}, 2017.

\bibitem[He et~al.(2016)He, Zhang, Ren, and Sun]{he2016deep}
Kaiming He, Xiangyu Zhang, Shaoqing Ren, and Jian Sun.
\newblock Deep residual learning for image recognition.
\newblock In \emph{Proceedings of the IEEE conference on computer vision and pattern recognition}, pages 770--778, 2016.

\bibitem[Ho et~al.(2020)Ho, Jain, and Abbeel]{ho2020denoising}
Jonathan Ho, Ajay Jain, and Pieter Abbeel.
\newblock Denoising diffusion probabilistic models.
\newblock \emph{Advances in neural information processing systems}, 33:\penalty0 6840--6851, 2020.

\bibitem[Hosseini and Poovendran(2018)]{hosseini2018semantic}
Hossein Hosseini and Radha Poovendran.
\newblock Semantic adversarial examples.
\newblock In \emph{Proceedings of the IEEE Conference on Computer Vision and Pattern Recognition Workshops}, pages 1614--1619, 2018.

\bibitem[Hu et~al.(2019)Hu, Yu, Guo, Chao, and Weinberger]{hu2019new}
Shengyuan Hu, Tao Yu, Chuan Guo, Wei-Lun Chao, and Kilian~Q Weinberger.
\newblock A new defense against adversarial images: Turning a weakness into a strength.
\newblock \emph{Advances in neural information processing systems}, 32, 2019.

\bibitem[Huang et~al.(2017)Huang, Liu, Van Der~Maaten, and Weinberger]{huang2017densely}
Gao Huang, Zhuang Liu, Laurens Van Der~Maaten, and Kilian~Q Weinberger.
\newblock Densely connected convolutional networks.
\newblock In \emph{Proceedings of the IEEE conference on computer vision and pattern recognition}, pages 4700--4708, 2017.

\bibitem[Jia et~al.(2022)Jia, Yin, Yao, Ding, Shen, Yang, and Ma]{jia2022adv}
Shuai Jia, Bangjie Yin, Taiping Yao, Shouhong Ding, Chunhua Shen, Xiaokang Yang, and Chao Ma.
\newblock Adv-attribute: Inconspicuous and transferable adversarial attack on face recognition.
\newblock \emph{Advances in Neural Information Processing Systems}, 35:\penalty0 34136--34147, 2022.

\bibitem[Laidlaw and Feizi(2019)]{laidlaw2019functionaladversarialattacks}
Cassidy Laidlaw and Soheil Feizi.
\newblock Functional adversarial attacks, 2019.

\bibitem[Laurent and Massart(2000)]{laurent2000adaptive}
Beatrice Laurent and Pascal Massart.
\newblock Adaptive estimation of a quadratic functional by model selection.
\newblock \emph{Annals of statistics}, pages 1302--1338, 2000.

\bibitem[Li et~al.(2023)Li, Su, Han, You, Huang, and Xu]{li2023adaptive}
Wenhao Li, Xiu Su, Yu Han, Shan You, Tao Huang, and Chang Xu.
\newblock Adaptive training meets progressive scaling: Elevating efficiency in diffusion models.
\newblock \emph{arXiv e-prints}, pages arXiv--2312, 2023.

\bibitem[Li et~al.(2024)Li, Zhang, Lan, and Jiang]{li2024transferable}
Wenyun Li, Zheng Zhang, Xiangyuan Lan, and Dongmei Jiang.
\newblock Transferable adversarial face attack with text controlled attribute.
\newblock \emph{arXiv preprint arXiv:2412.11735}, 2024.

\bibitem[Liang et~al.(2025)Liang, He, and Tan]{liang2025comprehensive}
Jian Liang, Ran He, and Tieniu Tan.
\newblock A comprehensive survey on test-time adaptation under distribution shifts.
\newblock \emph{International Journal of Computer Vision}, 133\penalty0 (1):\penalty0 31--64, 2025.

\bibitem[Liu et~al.(2023)Liu, Li, Wu, and Lee]{liu2023llava}
Haotian Liu, Chunyuan Li, Qingyang Wu, and Yong~Jae Lee.
\newblock Visual instruction tuning, 2023.

\bibitem[Liu et~al.(2021)Liu, Lin, Cao, Hu, Wei, Zhang, Lin, and Guo]{liu2021swin}
Ze Liu, Yutong Lin, Yue Cao, Han Hu, Yixuan Wei, Zheng Zhang, Stephen Lin, and Baining Guo.
\newblock Swin transformer: Hierarchical vision transformer using shifted windows.
\newblock In \emph{Proceedings of the IEEE/CVF international conference on computer vision}, pages 10012--10022, 2021.

\bibitem[Long et~al.(2022)Long, Zhang, Zeng, Gao, Liu, Zhang, and Song]{long2022frequency}
Yuyang Long, Qilong Zhang, Boheng Zeng, Lianli Gao, Xianglong Liu, Jian Zhang, and Jingkuan Song.
\newblock Frequency domain model augmentation for adversarial attack.
\newblock In \emph{European conference on computer vision}, pages 549--566. Springer, 2022.

\bibitem[Madry(2017)]{madry2017towards}
Aleksander Madry.
\newblock Towards deep learning models resistant to adversarial attacks.
\newblock \emph{arXiv preprint arXiv:1706.06083}, 2017.

\bibitem[Naseer et~al.(2020)Naseer, Khan, Hayat, Khan, and Porikli]{naseer2020self}
Muzammal Naseer, Salman Khan, Munawar Hayat, Fahad~Shahbaz Khan, and Fatih Porikli.
\newblock A self-supervised approach for adversarial robustness.
\newblock In \emph{Proceedings of the IEEE/CVF Conference on Computer Vision and Pattern Recognition}, pages 262--271, 2020.

\bibitem[Nichol et~al.(2021)Nichol, Dhariwal, Ramesh, Shyam, Mishkin, McGrew, Sutskever, and Chen]{nichol2021glide}
Alex Nichol, Prafulla Dhariwal, Aditya Ramesh, Pranav Shyam, Pamela Mishkin, Bob McGrew, Ilya Sutskever, and Mark Chen.
\newblock Glide: Towards photorealistic image generation and editing with text-guided diffusion models.
\newblock \emph{arXiv preprint arXiv:2112.10741}, 2021.

\bibitem[Nie et~al.(2022)Nie, Guo, Huang, Xiao, Vahdat, and Anandkumar]{nie2022diffusion}
Weili Nie, Brandon Guo, Yujia Huang, Chaowei Xiao, Arash Vahdat, and Anima Anandkumar.
\newblock Diffusion models for adversarial purification.
\newblock \emph{arXiv preprint arXiv:2205.07460}, 2022.

\bibitem[Oquab et~al.(2023)Oquab, Darcet, Moutakanni, Vo, Szafraniec, Khalidov, Fernandez, Haziza, Massa, El-Nouby, Howes, Huang, Xu, Sharma, Li, Galuba, Rabbat, Assran, Ballas, Synnaeve, Misra, Jegou, Mairal, Labatut, Joulin, and Bojanowski]{oquab2023dinov2}
Maxime Oquab, Timothée Darcet, Theo Moutakanni, Huy~V. Vo, Marc Szafraniec, Vasil Khalidov, Pierre Fernandez, Daniel Haziza, Francisco Massa, Alaaeldin El-Nouby, Russell Howes, Po-Yao Huang, Hu Xu, Vasu Sharma, Shang-Wen Li, Wojciech Galuba, Mike Rabbat, Mido Assran, Nicolas Ballas, Gabriel Synnaeve, Ishan Misra, Herve Jegou, Julien Mairal, Patrick Labatut, Armand Joulin, and Piotr Bojanowski.
\newblock Dinov2: Learning robust visual features without supervision, 2023.

\bibitem[Peebles and Xie(2023)]{peebles2023scalable}
William Peebles and Saining Xie.
\newblock Scalable diffusion models with transformers.
\newblock In \emph{Proceedings of the IEEE/CVF International Conference on Computer Vision}, pages 4195--4205, 2023.

\bibitem[Qiu et~al.(2020)Qiu, Xiao, Yang, Yan, Lee, and Li]{qiu2020semanticadv}
Haonan Qiu, Chaowei Xiao, Lei Yang, Xinchen Yan, Honglak Lee, and Bo Li.
\newblock Semanticadv: Generating adversarial examples via attribute-conditioned image editing.
\newblock In \emph{Computer Vision--ECCV 2020: 16th European Conference, Glasgow, UK, August 23--28, 2020, Proceedings, Part XIV 16}, pages 19--37. Springer, 2020.

\bibitem[Radford et~al.(2021)Radford, Kim, Hallacy, Ramesh, Goh, Agarwal, Sastry, Askell, Mishkin, Clark, et~al.]{radford2021learning}
Alec Radford, Jong~Wook Kim, Chris Hallacy, Aditya Ramesh, Gabriel Goh, Sandhini Agarwal, Girish Sastry, Amanda Askell, Pamela Mishkin, Jack Clark, et~al.
\newblock Learning transferable visual models from natural language supervision.
\newblock In \emph{International conference on machine learning}, pages 8748--8763. PMLR, 2021.

\bibitem[Rombach et~al.(2022)Rombach, Blattmann, Lorenz, Esser, and Ommer]{rombach2022high}
Robin Rombach, Andreas Blattmann, Dominik Lorenz, Patrick Esser, and Bj{\"o}rn Ommer.
\newblock High-resolution image synthesis with latent diffusion models.
\newblock In \emph{Proceedings of the IEEE/CVF conference on computer vision and pattern recognition}, pages 10684--10695, 2022.

\bibitem[Saharia et~al.(2022)Saharia, Chan, Saxena, Li, Whang, Denton, Ghasemipour, Gontijo~Lopes, Karagol~Ayan, Salimans, et~al.]{saharia2022photorealistic}
Chitwan Saharia, William Chan, Saurabh Saxena, Lala Li, Jay Whang, Emily~L Denton, Kamyar Ghasemipour, Raphael Gontijo~Lopes, Burcu Karagol~Ayan, Tim Salimans, et~al.
\newblock Photorealistic text-to-image diffusion models with deep language understanding.
\newblock \emph{Advances in neural information processing systems}, 35:\penalty0 36479--36494, 2022.

\bibitem[Sandler et~al.(2018)Sandler, Howard, Zhu, Zhmoginov, and Chen]{sandler2018mobilenetv2}
Mark Sandler, Andrew Howard, Menglong Zhu, Andrey Zhmoginov, and Liang-Chieh Chen.
\newblock Mobilenetv2: Inverted residuals and linear bottlenecks.
\newblock In \emph{Proceedings of the IEEE conference on computer vision and pattern recognition}, pages 4510--4520, 2018.

\bibitem[Shamsabadi et~al.(2020)Shamsabadi, Sanchez-Matilla, and Cavallaro]{shamsabadi2020colorfool}
Ali~Shahin Shamsabadi, Ricardo Sanchez-Matilla, and Andrea Cavallaro.
\newblock Colorfool: Semantic adversarial colorization.
\newblock In \emph{Proceedings of the IEEE/CVF conference on computer vision and pattern recognition}, pages 1151--1160, 2020.

\bibitem[Song and Ermon(2019)]{song2019generative}
Yang Song and Stefano Ermon.
\newblock Generative modeling by estimating gradients of the data distribution.
\newblock \emph{Advances in neural information processing systems}, 32, 2019.

\bibitem[Song et~al.(2020)Song, Sohl-Dickstein, Kingma, Kumar, Ermon, and Poole]{song2020score}
Yang Song, Jascha Sohl-Dickstein, Diederik~P Kingma, Abhishek Kumar, Stefano Ermon, and Ben Poole.
\newblock Score-based generative modeling through stochastic differential equations.
\newblock \emph{arXiv preprint arXiv:2011.13456}, 2020.

\bibitem[Szegedy(2013)]{szegedy2013intriguing}
C Szegedy.
\newblock Intriguing properties of neural networks.
\newblock \emph{arXiv preprint arXiv:1312.6199}, 2013.

\bibitem[Szegedy et~al.(2016)Szegedy, Vanhoucke, Ioffe, Shlens, and Wojna]{szegedy2016rethinking}
Christian Szegedy, Vincent Vanhoucke, Sergey Ioffe, Jon Shlens, and Zbigniew Wojna.
\newblock Rethinking the inception architecture for computer vision.
\newblock In \emph{Proceedings of the IEEE conference on computer vision and pattern recognition}, pages 2818--2826, 2016.

\bibitem[Tan and Le(2019)]{tan2019efficientnet}
Mingxing Tan and Quoc Le.
\newblock Efficientnet: Rethinking model scaling for convolutional neural networks.
\newblock In \emph{International conference on machine learning}, pages 6105--6114. PMLR, 2019.

\bibitem[Tian et~al.(2021)Tian, Zhou, Li, and Duan]{tian2021detecting}
Jinyu Tian, Jiantao Zhou, Yuanman Li, and Jia Duan.
\newblock Detecting adversarial examples from sensitivity inconsistency of spatial-transform domain.
\newblock In \emph{Proceedings of the AAAI conference on artificial intelligence}, pages 9877--9885, 2021.

\bibitem[Wang and He(2021)]{wang2021enhancing}
Xiaosen Wang and Kun He.
\newblock Enhancing the transferability of adversarial attacks through variance tuning.
\newblock In \emph{Proceedings of the IEEE/CVF conference on computer vision and pattern recognition}, pages 1924--1933, 2021.

\bibitem[Wang et~al.(2021)Wang, He, Wang, and He]{wang2021admix}
Xiaosen Wang, Xuanran He, Jingdong Wang, and Kun He.
\newblock Admix: Enhancing the transferability of adversarial attacks.
\newblock In \emph{Proceedings of the IEEE/CVF International Conference on Computer Vision}, pages 16158--16167, 2021.

\bibitem[Wang et~al.(2004)Wang, Bovik, Sheikh, and Simoncelli]{wang2004image}
Zhou Wang, Alan~C Bovik, Hamid~R Sheikh, and Eero~P Simoncelli.
\newblock Image quality assessment: from error visibility to structural similarity.
\newblock \emph{IEEE transactions on image processing}, 13\penalty0 (4):\penalty0 600--612, 2004.

\bibitem[Xiao et~al.(2018{\natexlab{a}})Xiao, Li, Zhu, He, Liu, and Song]{xiao2018generating}
Chaowei Xiao, Bo Li, Jun-Yan Zhu, Warren He, Mingyan Liu, and Dawn Song.
\newblock Generating adversarial examples with adversarial networks.
\newblock \emph{arXiv preprint arXiv:1801.02610}, 2018{\natexlab{a}}.

\bibitem[Xiao et~al.(2018{\natexlab{b}})Xiao, Zhu, Li, He, Liu, and Song]{xiao2018spatially}
Chaowei Xiao, Jun-Yan Zhu, Bo Li, Warren He, Mingyan Liu, and Dawn Song.
\newblock Spatially transformed adversarial examples.
\newblock \emph{arXiv preprint arXiv:1801.02612}, 2018{\natexlab{b}}.

\bibitem[Xie et~al.(2017)Xie, Wang, Zhang, Ren, and Yuille]{xie2017mitigating}
Cihang Xie, Jianyu Wang, Zhishuai Zhang, Zhou Ren, and Alan Yuille.
\newblock Mitigating adversarial effects through randomization.
\newblock \emph{arXiv preprint arXiv:1711.01991}, 2017.

\bibitem[Xiong et~al.(2022)Xiong, Lin, Zhang, Hopcroft, and He]{xiong2022stochastic}
Yifeng Xiong, Jiadong Lin, Min Zhang, John~E Hopcroft, and Kun He.
\newblock Stochastic variance reduced ensemble adversarial attack for boosting the adversarial transferability.
\newblock In \emph{Proceedings of the IEEE/CVF conference on computer vision and pattern recognition}, pages 14983--14992, 2022.

\bibitem[Xu et~al.(2023)Xu, Liu, Wu, Tong, Li, Ding, Tang, and Dong]{xu2023imagereward}
Jiazheng Xu, Xiao Liu, Yuchen Wu, Yuxuan Tong, Qinkai Li, Ming Ding, Jie Tang, and Yuxiao Dong.
\newblock Imagereward: Learning and evaluating human preferences for text-to-image generation.
\newblock \emph{Advances in Neural Information Processing Systems}, 36:\penalty0 15903--15935, 2023.

\bibitem[Yang et~al.(2024)Yang, Lin, Li, Zhao, Fan, Zhou, Wang, Liu, and Shen]{yang2024quantization}
Yulong Yang, Chenhao Lin, Qian Li, Zhengyu Zhao, Haoran Fan, Dawei Zhou, Nannan Wang, Tongliang Liu, and Chao Shen.
\newblock Quantization aware attack: Enhancing transferable adversarial attacks by model quantization.
\newblock \emph{IEEE Transactions on Information Forensics and Security}, 19:\penalty0 3265--3278, 2024.

\bibitem[Yuan et~al.(2022)Yuan, Zhang, Gao, Cheng, and Song]{yuan2022natural}
Shengming Yuan, Qilong Zhang, Lianli Gao, Yaya Cheng, and Jingkuan Song.
\newblock Natural color fool: Towards boosting black-box unrestricted attacks.
\newblock \emph{Advances in Neural Information Processing Systems}, 35:\penalty0 7546--7560, 2022.

\bibitem[Zhao et~al.(2020)Zhao, Liu, and Larson]{zhao2020adversarial}
Zhengyu Zhao, Zhuoran Liu, and Martha Larson.
\newblock Adversarial color enhancement: Generating unrestricted adversarial images by optimizing a color filter.
\newblock \emph{arXiv preprint arXiv:2002.01008}, 2020.

\end{thebibliography}
}
\clearpage
\setcounter{page}{1}
\maketitlesupplementary


\begin{table*}[t]
\centering
\scalebox{0.9}{
\begin{tabular}{ccccccccccc}
\hline
\multirow{2}{*}{\begin{tabular}[c]{@{}c@{}}Surrogate\\ Model\end{tabular}} & \multirow{2}{*}{Method} & \multicolumn{8}{c}{Attacked Models}                                                                                               & \multirow{2}{*}{\begin{tabular}[c]{@{}c@{}}Transfer\\ Avg \%\end{tabular}} \\ \cline{3-10}
                                                                           &                         & Mn-v2          & Inc-v3        & RN-50          & Den-161       & RN-152        & EF-b7         & Vit-b-32       & Swin-B         &                                                                            \\ \hline
-                                                                          & original                & 12.1           & 19.9          & 7.0            & 5.7           & 5.7           & 14.3          & 11.1           & 3.8            &                                                                            \\ \hline
\multirow{8}{*}{Swin-B}                                                    & SAE                     & 45.8           & 32.5          & 39.3           & 31.9          & 32.7          & 34.0          & 39.2           & 65.8*          & 36.5                                                                       \\
                                                                           & ReColorAdv              & 27.5           & 27.6          & 18.6           & 15.0          & 15.2          & 23.4          & 21.5           & 90.9*          & 21.3                                                                       \\
                                                                           & cAdv                    & 34.5           & 35.7          & 27.7           & 27.0          & 22.1          & 36.1          & 35.5           & 97.9*          & 31.2                                                                       \\
                                                                           & ColorFool               & 53.8           & 38.9          & 47.4           & 34.4          & 38.1          & 35.1          & 40.8           & 43.8*          & 41.2                                                                       \\
                                                                           & NCF                     & 56.1           & 39.8          & 51.1           & 38.9          & 43.3          & 38.8          & 49.1           & 64.4*          & 45.3                                                                       \\
                                                                           & ACA                     & 66.1           & 66.5          & 63.6           & 61.8          & 61.5          & 62.6          & \textbf{61.8}  & 81.4*          & 63.4                                                                       \\
                                                                           & DiffAttack              & 60.3           & 57.6          & 57.5           & 57.6          & 55.0          & 55.3          & 44.9           & \textbf{91.3*} & 55.5                                                                       \\
                                                                           & ours                    & \textbf{71.0}  & \textbf{67.1} & \textbf{69.8}  & \textbf{62.5} & \textbf{61.7} & \textbf{64.5} & 51.0           & 89.9*          & \textbf{63.9}                                                              \\ \hline
\end{tabular}
}
\caption{Performance comparison of adversarial transferability on normally trained CNNs and ViTs. The adversarial examples are generated by Swin-B, respectively. "*" indicates white-box attack settings.}
\label{table_swinb}
\end{table*}

\section{Proof for Proposition}
\label{sec:proof}
The denoising steps in the Diffusion Model are discrete, making theoretical analysis highly challenging. \cite{song2020score} established a connection between the DDPM~\cite{ho2020denoising} process and stochastic differential equations (SDEs~\cite{song2019generative}), showing that DDPM can be expressed as a specific form of SDE. We map the discrete timesteps $k\in\{0,1,\cdots,T\}$ to a continuous time variable $t\in[0,1]$ by normalizing the discrete index into a unit interval. Accordingly, any function originally defined on discrete steps, such as the noise schedule $\beta(k)$, is redefined as a continuous function $\beta(t)$. As $T\to\infty$, DDPM gradually approximates a continuous process. The forward DDPM can be expressed by:
\begin{equation}
dx_t=-\frac{1}{2}\beta(t)x_tdt+\sqrt{\beta(t)}dw_t,
\label{eq:sde_forward}
\end{equation}
The reverse SDE can be expressed by:
\begin{equation}
dx_t=[-\frac{1}{2}\beta(t)x_t-\beta(t)\nabla_x\log p_t(x)]dt+\sqrt{\beta(t)}dw_t,
\label{eq:sde_reverse}
\end{equation}

\noindent
\textbf{Proposition}. \textit{Let $z_0(t^*)$ be the denoised latent variable generated from the noisy latent $z_{t^*}$ via the reverse SDE process. Assume the normalized noise prediction error satisfies $||-\frac{1}{\sqrt{1-\bar{\alpha}_t}}\epsilon_\theta(z_t,t)||_2^2\leq C$ for all $z_t$, $t\in[0,1]$, and let $\delta\in(0,1)$. Then, with probability at least $(1-\delta)$, }
\begin{equation}
\begin{aligned}
&\|z-z_0(t^*)\|_2^2 \\ 
\leq \ & \sigma^2(t^*)(C\sigma^2(t^*)+d_z+2\sqrt{-d_z\log\delta}-2\log\delta),
\end{aligned}
\label{eq:proposition2}
\end{equation}
\textit{where $z=E(x)$ is the VAE-encoded latent, $z_0(t^*)$ is the denoising result from $z_{t^*}$ and $d_z$ is the latent dimension of $z$.}

\noindent
\textbf{Proof}. 
We first prove the proposition holds in DDPM. We integrate from $t^*$ to $0$ using Eq.\ref{eq:sde_reverse}:
\begin{equation}
\begin{aligned}
&x_0-x_{t^*}\\
=\ &\underbrace{\int_{t^*}^0[-\frac{1}{2}\beta(t)x_t-\beta(t)\nabla_x\log p_t(x)]dt}_A+\underbrace{\int_{t^*}^0\sqrt{\beta(t)}dw_t}_B,
\end{aligned}
\end{equation}
using the identity $\nabla_{x_t}\log p_t(x_t)=-\frac{\epsilon_\theta(x_t,t)}{\sqrt{1-\bar{\alpha_t}}}$, we rewrite:
\begin{equation}
A=\int_{t^*}^0\beta(t)(-\frac{1}{2}x_t+\frac{\epsilon_\theta(x_t,t)}{\sqrt{1-\bar{\alpha_t}}})dt,
\end{equation}
by hypothesis $\|\frac{\epsilon_\theta}{\sqrt{1-\bar{\alpha}_t}}\|_2^2\leq C$, and under the forward process $x_{t^*}\sim\mathcal{N}(0,\sigma^2(t^*)I)$, we bound:
\begin{equation}
\|A\|_2^2\leq(\int_{t^*}^0\beta(t)(\frac{\sqrt{d\sigma^2(t^*)}}{2}+\sqrt{C})dt)^2\leq C\sigma^4(t^*).
\end{equation}
The Itô integral $B\sim\mathcal{N}(0,\sigma^2(t^*)I)$. By \cite{laurent2000adaptive}, for $\delta\in(0,1)$:
\begin{equation}
\mathbb{P}(\frac{\|B\|_2^2}{\sigma^2(t^*)}\geq d+2\sqrt{-d\log\delta}-2\log\delta)\leq \delta.
\end{equation}
With probability $1-\delta$:
\begin{equation}
\begin{aligned}
&\|x_0-x_{t^*}\|_2^2\\
\leq \ &C\sigma^4(t^*)+\sigma^2(t^*)(d+2\sqrt{-d\log\delta}-2\log\delta).
\end{aligned}
\end{equation}
This establishes the proposition for DDPM.

Now we generalization the proposition to latent diffusion model.
Latent diffusion model maps $x$ to $z=E(x)$. If the encoder $E$ ensures $z\sim\mathcal{N}(0,I)$, then the latent diffusion SDE mirrors DDPM:
\begin{equation}
d_{z_t}=-\frac{1}{2}\beta(t)z_tdt+\sqrt{\beta(t)}dw_t.
\end{equation}
The reverse SDE is structurally identical:
\begin{equation}
d_{z_t}=[-\frac{1}{2}\beta(t)z_tdt-\beta(t)\nabla_{z_t}\log p_t(z_t)]dt+\sqrt{\beta(t)}dw_t.
\end{equation}
The latent score satisfies $\nabla_{z_t}\log p_t(z_t)=-\frac{\epsilon_\theta(z_t,t)}{\sqrt{1-\bar{\alpha}_t}}$, analogous to DDPM. By hypothesis, $\|-\frac{\epsilon_\theta(z_t,t)}{\sqrt{1-\bar{\alpha}_t}}\|_2^2\leq C$, preserving the deterministic term bound $\|A\|_2^2\leq C\sigma^4(t^*)$.
If the latent dimension $d \ll d_x$ replaces $d$, then the stochastic term bound becomes:
\begin{equation}
\|B\|_2^2\leq\sigma^2(t^*)(d_z+2\sqrt{-d_z\log\delta}-2\log\delta).
\end{equation}
Let $z_0(t^*)$ denote the denoised latent. The VAE decoder $D$ maps $z_0(t^*)$ to $\hat{x}=D(z_0(t^*))$. If $D$ is $K$-Lipschitz:
\begin{equation}
\|D(z_0(t^*))-D(z_{t^*})\|\leq K\|z_0(t^*)-z_{t^*}\|_2+\epsilon_{\text{recon}},
\end{equation}
where $\epsilon_\text{recon}$ is the VAE’s reconstruction error. Substituting the latent-space bound:
\begin{equation}
\begin{aligned}
&\|z_0(t^*)-z_{t^*}\|_2^2\\
\leq\ &\sigma^2(t^*)(C\sigma^2(t^*)+d_z+2\sqrt{-d_z\log\delta}-2\log\delta).
\end{aligned}
\end{equation}

In summary, this proposition can be extended to the latent diffusion model under the following conditions: 
\begin{itemize}
\item \textbf{Consistent SDE Dynamics}: The latent diffusion process follows the same forward and reverse SDE structure as DDPM, specifically
\begin{equation}
d_{z_t}=-\frac{1}{2}z_tdt+\sqrt{\beta(t)}dw_t,
\end{equation}
ensuring identical noise accumulation $\sigma^2(t^*)=1-\bar{\alpha}_{t^*}$.

\item \textbf{Score Function Boundedness}: The normalized noise prediction term , which approximates the latent score $\nabla_{z_t}\log p_t(z_t)$, satisfies $\|-\epsilon_\theta/\sqrt{1-\bar{\alpha}_t}\|_2^2\leq C$, inherited from the training objective of denoising score matching.

\item \textbf{Gaussian Latent Alignment}: The VAE encoder enforces $z\sim\mathcal{N}(0,I)$, preserving the Gaussianity of the latent forward process and ensuring the stochastic term $\int\sqrt{\beta(t)}dw_t$ follows a scaled $\chi^2$-distribution.

\item \textbf{Dimensional Adaptation}: The dimension $d$ in the original bound is replaced by the latent dimension $d_z$, which is typically orders of magnitude smaller than the pixel space, leading to tighter bounds due to reduced dimensionality.

\item \textbf{Controlled Decoding Error}: The VAE decoder $D$ is Lipschitz continuous, propagating the latent gap $\|z-z_0(t^*)\|_2$ to the pixel space with bounded amplification.

\end{itemize}

\begin{figure}[t]
\centering
\scalebox{1.0}{
  \includegraphics[width=1\linewidth]{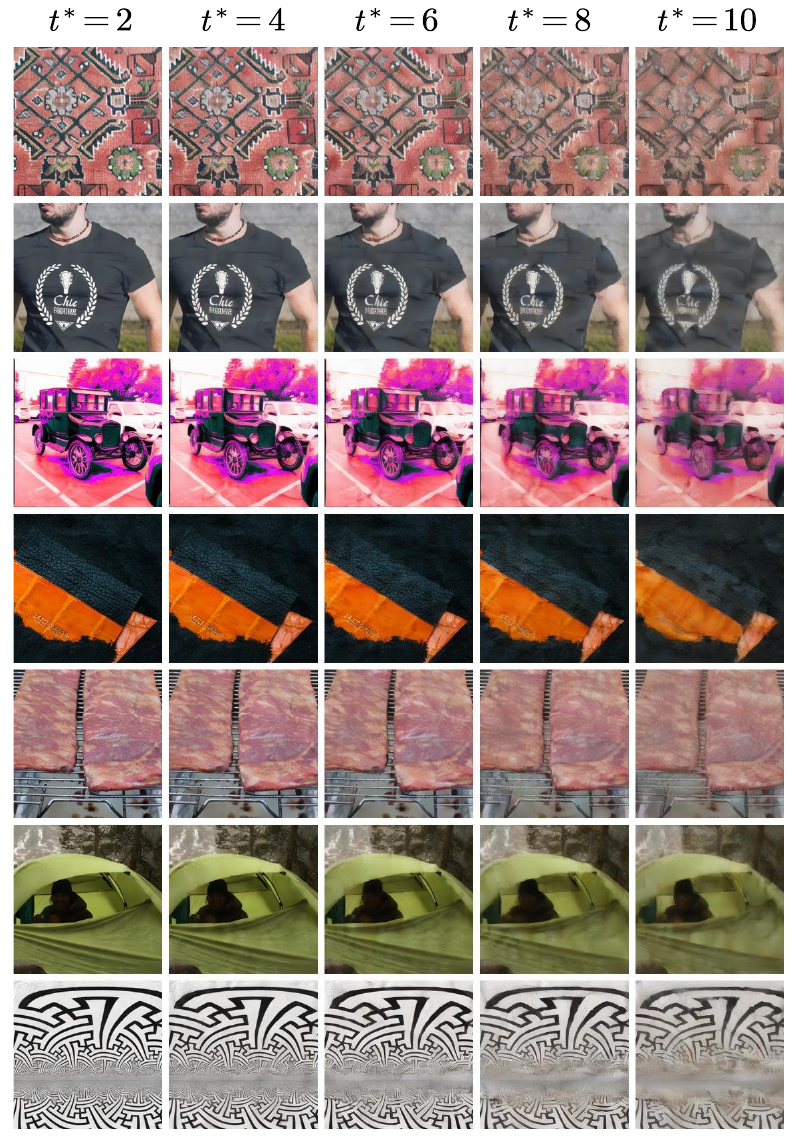} 
  }
   \caption{We visualize adversarial examples generated under the same settings with different $t^*$ values. As $t^*$ increases, the overall image content remains unchanged, but the quality degrades due to the excessive injection of adversarial perturbations. (We selected some more noticeable examples to compare the effects of different $t^*$ values.)}
   \label{fig:supp_trade_off}
\end{figure}

\section{More Experiments}
\label{sec:more_exp}
\textbf{More Comparisons}. In the experiments of Section \ref{subsec:4-2}, we also used Swin-B as the surrogate model, as shown in Table \ref{table_swinb}. 

\noindent
\textbf{More Trade-off Showcase}. In Section \ref{subsec:exp_trade_off}, we demonstrated that selecting $t^*$ requires balancing attack transferability and the similarity between the generated and original images. Here, we present a comparison illustrating how different $t^*$ values affect the generated images, as shown in the Figure \ref{fig:supp_trade_off}.

\end{document}